\documentclass[11pt,a4paper]{article}
\usepackage{times,latexsym}
\usepackage{url}
\usepackage[T1]{fontenc}
\usepackage[acceptedWithA]{tacl2018v2}

\usepackage{xspace,mfirstuc,tabulary}

\newif\iftaclinstructions
\taclinstructionsfalse % AUTHORS: do NOT set this to true
\iftaclinstructions
\renewcommand{\confidential}{}
\renewcommand{\anonsubtext}{(No author info supplied here, for consistency with
TACL-submission anonymization requirements)}
\newcommand{\instr}
\fi

\iftaclpubformat % this "if" is set by the choice of options

\else

\fi

\usepackage{times}
\usepackage{latexsym}
\usepackage{amsfonts}
\usepackage{comment}
\usepackage{tikz}
\usepackage{pgfplots}
\usetikzlibrary{shapes, positioning, decorations.pathreplacing, shapes.multipart}
\usepackage{amsmath}
\usepackage{todonotes}
\usepackage{algorithm}
\usepackage{multirow}
\usepackage{subfig}
\usepackage[noend]{algpseudocode}
\usepackage{booktabs}
\usepackage{url}
\usepackage{mymathdef}
\usepackage[T1]{fontenc}
\usepackage[utf8]{inputenc}
\usepackage{CJKutf8}
 \usepackage{threeparttable}
\usepackage{tikz}
\usepackage{tikz-dependency}
\usepackage{graphicx}
\usepackage{url}

\date{}
\author{
 Hang Yan,\ Xipeng Qiu\Thanks{Corresponding author.}\ ,\ Xuanjing Huang \\
 School of Computer Science, Fudan University, China \\
 Shanghai Key Laboratory of Intelligent Information Processing, Fudan University, China\\
  {\sf \{hyan19, xpqiu, xjhuang\}@fudan.edu.cn} \\
}

\begin{document}
\title{A Graph-based Model for Joint Chinese Word Segmentation and Dependency Parsing}
% \author{
% Hang Yan, Xipeng Qiu, Xuanjing Huang
% %Michael~Shell,~\IEEEmembership{Member,~IEEE,}
% %John~Doe,~\IEEEmembership{Fellow,~OSA,}
% %and~Jane~Doe,~\IEEEmembership{Life~Fellow,~IEEE}% <-this % stops a space
% \thanks{
% H. Yan, X. Qiu and X. Huang are with the School of Computer Science and the Shanghai Key Laboratory of Intelligent Information Processing, Fudan University, Shanghai 200433, China (e-mail:
% 11300720199@fudan.edu.cn;xpqiu@fudan.edu.cn;xjhuang@fudan.edu.cn).
% }% <-this % stops a space
% %\thanks{J. Doe and J. Doe are with Anonymous University.}% <-this % stops a space
% %\thanks{Manuscript received April 19, 2005; revised August 26, 2015.}
% }

\maketitle

\begin{abstract}

Chinese word segmentation and dependency parsing are two fundamental tasks for Chinese natural language processing. The dependency parsing is defined on word-level. Therefore word segmentation is the precondition of dependency parsing, which makes dependency parsing suffer from error propagation and unable to directly make use of the character-level pre-trained language model (such as BERT). In this paper, we propose a graph-based model to integrate Chinese word segmentation and dependency parsing.
Different from previous transition-based joint models, our proposed model is more concise, which results in fewer efforts of feature engineering. Our graph-based joint model achieves better performance than previous joint models and state-of-the-art results in both Chinese word segmentation and dependency parsing.
Besides, when BERT is combined, our model can substantially reduce the performance gap of dependency parsing between joint models and gold-segmented word-based models. Our code is publicly available at \url{https://github.com/fastnlp/JointCwsParser}.

\end{abstract}

\begin{CJK*}{UTF8}{gbsn}
\section{Introduction}

Unlike English, Chinese sentences consist of continuous characters and lack obvious boundaries between Chinese words. Since words are usually regarded as the minimum semantic units, therefore Chinese word segmentation (CWS) becomes a preliminary pre-process step for downstream Chinese natural language processing (NLP). For example, the fundamental NLP task, dependency parsing, is usually defined on word-level. To parse a Chinese sentence, the process is usually the following pipeline way: word segmentation, Part-of-Speech (POS) tagging and dependency parsing.

However, the pipeline way always suffers from the following limitations:

(1) Error Propagation.
In the pipeline way, once some words are wrongly segmented, the subsequent POS tagging and parsing will also make mistakes. As a result, pipeline models achieve dependency scores of around $75\sim 80\%$~\cite{DBLP:conf/acl/KuritaKK17}.

(2) Knowledge Sharing. These three tasks (word segmentation, POS tagging, and dependency parsing) are strongly related. The criterion of Chinese word segmentation also depends on the grammatical role in a sentence.
Therefore, the knowledge learned from these three tasks can be shared. The knowledge of one task can help others.
However, the pipeline way separately trains three models, each for a single task, and cannot fully exploit the shared knowledge among the three tasks.

\begin{figure*}[t]
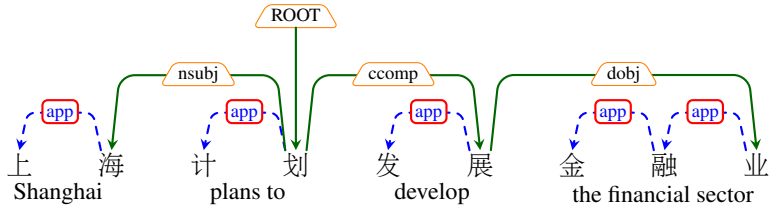

    \centering\small
\depstyle{app}{edge style = {thick, blue,dashed},
label style = {thick, draw=red, text=blue, fill=white}}
\depstyle{dep}{edge theme = grassy, label style={draw=orange,trapezium}}

\tikzstyle{en}=[font=\footnotesize\selectfont,text=black]
\tikzstyle{mypos}=[font=\footnotesize,text=blue!60!black]

\begin{dependency}%[theme = simple]
   \begin{deptext}[column sep=2em]
      上 \& 海 \& 计 \& 划  \& 发 \& 展\& 金\& 融 \& 业  \\
   \end{deptext}
   \node (shanghai) [en,below of = \wordref{1}{2}, node distance=1em, xshift=-2em] {Shanghai};
 %  \node (pos2) [mypos,below of = \wordref{1}{2}, node distance=3em] {NR};
%   \draw [->,thick, red] (\wordref{1}{2}) -- (pos2);

   \node (plan) [en,below of = \wordref{1}{4}, node distance=1em, xshift=-1.8em] {plans to};
%   \node (pos4) [mypos,below of = \wordref{1}{4}, node distance=3em] {VV};
%   \draw [->,thick, red] (\wordref{1}{4}) -- (pos4);

   \node (VV) [en,below of = \wordref{1}{6}, node distance=1em, xshift=-1.8em] {develop};
%   \node (pos6) [mypos,below of = \wordref{1}{6}, node distance=3em] {VV};
%   \draw [->,thick, red] (\wordref{1}{6}) -- (pos6);

   \node (VV) [en,below of = \wordref{1}{9}, node distance=1em, xshift=-3.5em] {the financial sector};
%   \node (pos9) [mypos,below of = \wordref{1}{9}, node distance=3em] {NN};
%   \draw [->,thick, red] (\wordref{1}{9}) -- (pos9);

   \deproot[dep]{4}{ROOT}
   \depedge[app]{4}{3}{app}
   \depedge[app]{2}{1}{app}
   \depedge[app]{6}{5}{app}
   \depedge[app]{8}{7}{app}
   \depedge[app]{9}{8}{app}

   %\depedge[edge start x offset=-6pt]{2}{5}{nsubj}

   \depedge[dep]{4}{2}{nsubj}
   \depedge[dep]{4}{6}{ccomp}
   \depedge[dep,edge unit distance=2.1ex]{6}{9}{dobj}
   %\depedge[]{7}{6}{ATT}
\end{dependency}
    \caption{The unified framework for joint Chinese word segmentation and dependency parsing. The green arc indicates the word-level dependency relation. The dashed blue arc with ``app'' indicates its connected characters belong to a word.}\label{fig:char-parsing}
\end{figure*}

A traditional solution to this error propagation
problem is to use joint models~\cite{DBLP:conf/acl/HatoriMMT12,DBLP:conf/acl/ZhangZCL14,DBLP:conf/acl/KuritaKK17}.
These previous joint models mainly adopted a transition-based parsing framework to integrate the word
segmentation, POS tagging and dependency parsing.
Based on the standard sequential shift-reduce transitions, they design some extra actions for word segmentation and POS tagging. Although these joint models achieved better performance than the pipeline model, they still suffer from two limitations:

(1) The first is the huge search space. Compared to the word-level transition parsing, the character-level transition parsing has longer sequence of actions. The search space is huge. Therefore, it is hard to find the best transition sequence exactly. Usually, the approximate strategies like greedy search or beam search are adopted in practice. However, the approximate strategies do not, in general, produce an optimal solution. Although exact searching is possible within $O(n^3)$ complexity~\cite{DBLP:conf/emnlp/ShiHL17}. Due to their complexity, these models just focus on unlabeled dependency parsing, rather than labeled dependency parsing.

(2) The second is the feature engineering. These transition-based joint models rely on a detailed handcrafted feature. Although \citet{DBLP:conf/acl/KuritaKK17} introduced neural models to reduce partial efforts of feature engineering, they still require hard work on how to design and compose the word-based features from the stack and the character-based features from the buffer.

Recently, the graph-based models have made significant progress for dependency parsing~\cite{DBLP:journals/tacl/KiperwasserG16,DBLP:conf/iclr/DozatM17}, which fully exploit the ability of the bidirectional long short-term memory network (BiLSTM)~\cite{DBLP:journals/neco/HochreiterS97} and attention mechanism~\cite{DBLP:journals/corr/BahdanauCB14} to capture the interactions of words in a sentence. Different from the transition-based models, the graph-based models
assign a score or probability to each possible arc and then construct a maximum spanning tree (MST) from these weighted arcs.

In this paper, we propose a joint model for Chinese word segmentation and dependency parsing, which integrates these two tasks into a unified graph-based parsing framework. Since the segmentation is a character-level task and dependency parsing is a word-level task, we first formulate these two tasks into character-level graph-based parsing framework.
In detail, our model contains (1) a deep neural network encoder, which can capture the long-term contextual features for each character, it can be multi-layer BiLSTM or pre-trained BERT, (2) a biaffine attentional scorer~\cite{DBLP:conf/iclr/DozatM17} , which unifies segmentation and dependency relations on character-level.  Besides, unlike the previous joint models~~\cite{DBLP:conf/acl/HatoriMMT12,DBLP:conf/acl/ZhangZCL14,DBLP:conf/acl/KuritaKK17},
our joint model does not depend on the POS tagging task.

%distinguish the head and dependent characters by mapping them into two different vector spaces, and
In the experiments on three popular datasets, we obtain state-of-the-art performance on Chinese word segmentation and dependency parsing.

In this paper, we claim four contributions as the following:
\begin{itemize*}
    \item To the best of our knowledge, this is the first graph-based method to integrate Chinese word segmentation and dependency parsing both in the training phase and decoding phase. The proposed model is very concise and easily implemented.
    \item Compared to the previous transition-based joint models, our proposed model is a graph-based model, which results in fewer efforts of feature engineering. Besides, our model can deal with the labeled dependency parsing task, which is not easy for the transition-based joint models.
    \item In experiments on datasets CTB-5, CTB-7 and CTB-9, our model achieves the state-of-the-art scores in joint Chinese word segmentation and dependency parsing, even without the POS information.
    \item As an added bonus, our proposed model can directly utilize the pre-trained language model BERT~\cite{DBLP:conf/naacl/DevlinCLT19} to boost performance significantly. The performance of many NLP tasks can be significantly enhanced when BERT was combined \cite{DBLP:conf/naacl/SunHQ19,DBLP:conf/acl/ZhongLWQH19}. However, for Chinese, BERT is based on Chinese characters, while dependency parsing is conducted in the word-level. We cannot directly utilize BERT to enhance the word-level Chinese dependency parsing models.  Nevertheless, by using the our proposed model, we can exploit BERT to implement Chinese word segmentation and dependency parsing jointly.
\end{itemize*}

\section{Related Work}

%Although neural network models for separate word segmentation and POS tagging have been intensively investigated, only a few works have been aware of the neural joint model of the two tasks.

To reduce the problem of error propagation and improve the low-level tasks by incorporating the knowledge from the high-level tasks, many successful joint methods have been proposed to simultaneously solve related tasks, which can be categorized into three types.

\subsection{Joint Segmentation and POS tagging}

Since the segmentation is a character-level task and POS tagging is a word-level task, an intuitive idea is to transfer both the tasks into character-level and incorporate them in a uniform framework.

A popular method is to assign a cross-tag to each character~\cite{DBLP:conf/emnlp/NgL04}. The cross-tag is composed of a word boundary part and a POS part,
e.g., ``B-NN'' refers to the first character in a word
with POS tag ``NN''. Thus, the joint CWS and POS tagging can be regarded as a sequence labeling problem. Following this work, \citet{DBLP:conf/emnlp/ZhengCX13,DBLP:conf/ijcai/ChenQH17,DBLP:conf/ijcnlp/ShaoHTN17} utilized neural models to alleviate the efforts of feature engineering.

Another line of the joint segmentation and POS
tagging method is the transition-based method~\cite{DBLP:conf/acl/ZhangC08,DBLP:conf/emnlp/ZhangC10a}, in which the joint decoding process is regarded as a sequence of action predictions.
%Zhang and Clark (2010) improve this model by using both character and word-based decoding.
\citet{DBLP:journals/taslp/ZhangYF18}  used a simple yet effective sequence-to-sequence neural model to improve the performance of the transition-based method.

\subsection{Joint POS tagging and Dependency Parsing}

Since the POS tagging task and dependency parsing task are word-level tasks, it is more natural to combine them into a joint model.

\citet{DBLP:conf/acl/HatoriMMT12} proposed a transition-based joint POS tagging and dependency parsing model and showed that the joint approach improved the accuracies of these two tasks. \citet{DBLP:journals/taslp/YangZLSYF18} extended this model by neural models to alleviate the efforts of feature engineering.

\citet{DBLP:conf/emnlp/LiZCLCL11} utilized the graph-based model to jointly optimize POS tagging and dependency parsing in a unique model. They also proposed an effective POS tag pruning method which could greatly improve the decoding efficiency.

By combining the lexicality and syntax into a unified framework, joining POS tagging and dependency parsing can improve both tagging and parsing performance over independent modeling significantly.

\subsection{Joint Segmentation, POS tagging and Dependency Parsing}

Compared to the above two kinds of joint tasks, it is non-trivial to incorporate all the three tasks into a joint model.

\citet{DBLP:conf/acl/HatoriMMT12} first proposed a transition-based joint model for Chinese word
segmentation, POS tagging and dependency parsing, which stated that dependency information improved the performances of word segmentation and POS tagging. \citet{DBLP:conf/acl/ZhangZCL14}
expanded this work by using intra-character structures
of words and found the intra-character dependencies were helpful in word segmentation and POS tagging. \citet{DBLP:conf/naacl/ZhangLBD15} proposed joint segmentation, POS tagging and dependency re-ranking system. This system required a base parser to generate some candidate parsing results.
 %Zhu et al. (2015) propose the re-ranking system of parsing results with a recursive convolutional neural network.
\citet{DBLP:conf/acl/KuritaKK17} followed the work of~\citet{DBLP:conf/acl/HatoriMMT12,DBLP:conf/acl/ZhangZCL14} and used the bidirectional LSTM (BiLSTM) to extract features with n-gram character string embeddings as input.

A related work is the full character-level neural dependency parser~\cite{DBLP:conf/aaai/LiZJZ18}, but it focuses on character-level parsing without considering the word segmentation and word-level POS tagging and parsing. Although a heuristic method could transform the character-level parsing results to word-level, the transform strategy is tedious and the result is also worse than other joint models.

Besides, there are some joint models for constituency parsing. \citet{DBLP:conf/emnlp/QianL12} proposed a joint inference model for word segmentation, POS tagging and constituency parsing. However, their model did not train three tasks jointly and suffered from the
decoding complexity due to the large combined search space. \citet{DBLP:conf/acl/WangZX13} firstly segmented a Chinese
sentence into a word lattice, and then predicted the POS tags and parsed tree based on the word lattice. A dual decomposition method was employed to encourage the tagger and parser to predict agreed structures.

The above methods show that syntactic parsing can provide useful feedback to word segmentation and POS tagging and the joint inference leads to improvements in all three sub-tasks.
Moreover, there is no related work on joint Chinese word segmentation and dependency parsing, without POS tagging.

\section{Proposed Model}

Previous joint methods are mainly based on the transition-based model, which modify the standard ``shift-reduce'' operations by adding some extra operations, such as ``app'' and ``tag''. Different from previous methods, we integrate word segmentation and dependency parsing into a graph-based parsing framework, which is simpler and easily implemented.

\begin{figure}[t]
    \centering
    \includegraphics[width=0.5\textwidth]{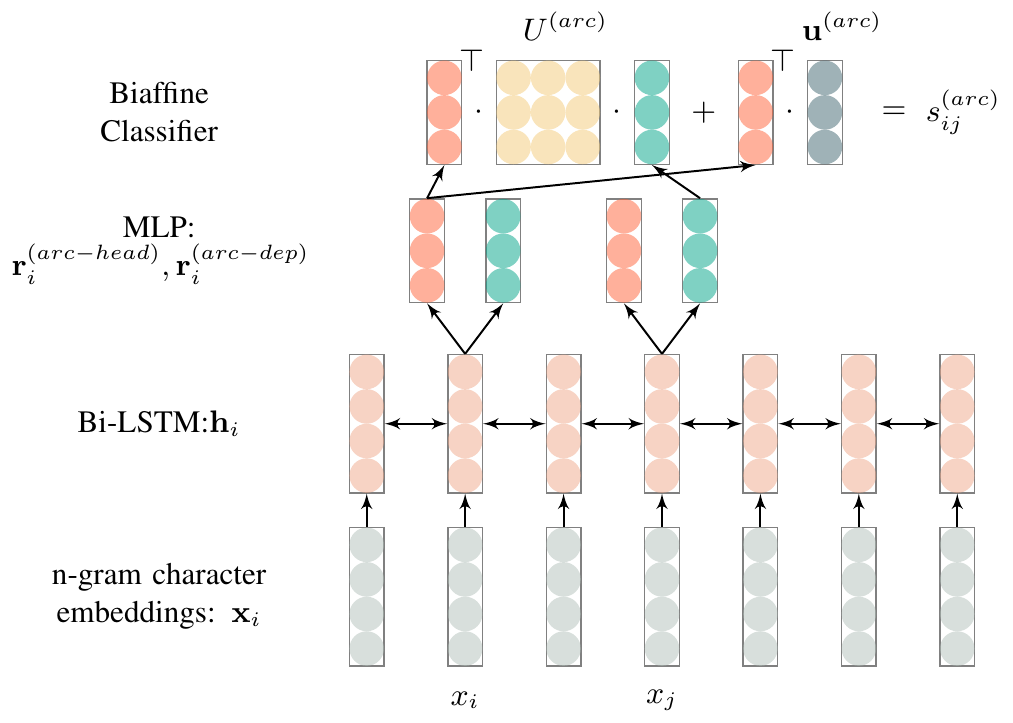}
    \caption{Proposed joint model when the encoder layer is BiLSTM. For simplicity, we omit the prediction of arc label, which uses a different biaffine classifier.}\label{fig:arch}
  \end{figure}

Firstly, we transform the word segmentation to a special arc prediction problem. For example, a Chinese word ``金融业 (financial sector)'' has two intra-word dependent arcs: ``金$\leftarrow$融'' and ``融$\leftarrow$业''. Both intra-word dependent arcs have the label ``app''. In this work, all characters in a word (excluding the last character) depend on their latter character, as the ``金融业 (financial sector)'' in Fig. \ref{fig:char-parsing}. Character-based dependency parsing arc has also been used in~\citet{DBLP:conf/acl/HatoriMMT12,DBLP:conf/acl/ZhangZCL14}, but their models were transition-based.

Secondly, we transform the word-level dependency arcs to character-level dependency arcs.
Assuming that there is a dependency arc between words $w_1={x_{i:j}}$ and $w_2={x_{u:v}}$, where $x_{i:j}$ denotes the continuous characters from $i$ to $j$ in a sentence, we make this arc to connect the last characters $x_j$ and $x_v$ of each word.
For example, the arc ``发展(develop)$\rightarrow$金融业 (financial sector)'' is translated to ``展$\rightarrow$业''. Fig. \ref{fig:char-parsing} illustrate the framework for joint Chinese word segmentation and dependency parsing.

Thus, we can use a graph-based parsing model to conduct these two tasks. Our model contains two main components: (1) a deep neural network encoder to extract the contextual features, which converts discrete characters into dense vectors, (2) a biaffine attentional scorer~\citep{DBLP:conf/iclr/DozatM17} which takes the hidden vectors for the given character pair
as input and predict a label score vector.

 Fig. \ref{fig:arch} illustrates the model structure for joint Chinese word segmentation and dependency parsing. The detailed description is as follows.

\subsection{Encoding Layer}

The encoding layer is responsible for converting discrete characters into contextualized dense representations. In this paper, we tried two different kinds of encode layers. The first one is multi-layer BiLSTM, the second one is the pre-trained language model BERT~\cite{DBLP:conf/naacl/DevlinCLT19} which is based on self-attention.

\subsubsection{BiLSTM-based Encoding Layer}

Given a character sequence $X = \{x_1, \dots, x_N\}$, in neural models, the first step is to map discrete language symbols into distributed embedding space. Formally, each character $x_i$ is mapped as $\be_i \in \mathbb{R}^{d_e} \subset \bE$, where $d_e$ is a hyper-parameter indicating the size of character embedding, and $\bE$ is the embedding matrix.
Character bigrams and trigrams have been shown
highly effective for Chinese word segmentation and POS tagging in previous studies~\cite{DBLP:conf/acl/PeiGC14,DBLP:conf/emnlp/ChenQZLH15,DBLP:conf/ijcnlp/ShaoHTN17,DBLP:journals/taslp/ZhangYF18}. Following their settings, we combine the character bigram and trigram to enhance the representation of each character.
The final character representation of $x_i$ is
given by $\be_i = \be_{x_{i}}\oplus \be_{x_{i}x_{i+1}} \oplus \be_{x_{i}x_{i+1}x_{i+2}}$, where $\be$ denotes the embedding for unigram, bigram and trigram, and $\oplus$ is the concatenation operator.

To capture the long-term contextual information, we use a deep bi-directional LSTM (BiLSTM)~\cite{DBLP:journals/neco/HochreiterS97} to incorporate information from both sides of a sequence, which is a prevalent choice in recent research for NLP tasks.

The hidden state of LSTM for the $i$-th character is
\begin{equation}
\bh_i = \BiLSTM(\be_{i}, \overrightarrow{\bh}_{i-1}, \overleftarrow{\bh}_{i+1}, \theta),\label{eq:LSTM}
\end{equation}
where $\overrightarrow{\bh}_i$ and $\overleftarrow{\bh}_i$ are the hidden states at position $i$ of the forward and backward LSTMs respectively, and $\theta$ denotes all the parameters in BiLSTM layer.

\subsubsection{BERT-based Encoding Layer}

Other than using BiLSTM as the encoder layer, pre-trained BERT can also been used as the encoding layer~\cite{DBLP:conf/naacl/DevlinCLT19,DBLP:journals/corr/abs-1906-08101}. The input of BERT is the character sequence $X = \{x_1, \dots, x_N\}$, the output of the last layer of BERT is used as the representation of characters. More details on the structure of BERT can be found in ~\citet{DBLP:conf/naacl/DevlinCLT19}.

\subsection{Biaffine Layer}

To predict the relations of each character pair, we employ the biaffine attention mechanism~\cite{DBLP:conf/iclr/DozatM17} to score their probability on the top of
 encoding layers. According to~\citet{DBLP:conf/iclr/DozatM17}, biaffine attention is more effectively capable of measuring the relationship between two elementary units.

%Note that
%biaffine attention is a natural extension of bilinear attention
%(Luong, Pham, and Manning 2015) which is widely used in
%neural machine translation (NM

\subsubsection{Unlabeled Arc Prediction} \label{Unlabeled Arc Prediction}
For the pair of the $i$-th and $j$-th characters, we first take the output of the encoding layer $\bh_i$ and $\bh_j$, then feed them into an extension of
bilinear transformation called a \textit{biaffine function}
to obtain the score for an arc from $x_i$ (head) to $x_j$ (dependent).

\begin{align}
    \br_i^{(arc-head)} &=\mathrm{MLP}^{(arc-head)}(\bh_i), \\
    \br_j^{(arc-dep)} &= \mathrm{MLP}^{(arc-dep)} (\bh_j), \\
    s_{ij}^{(arc)} &= \br_i^{(arc-head)} U^{(arc)} \br_j^{(arc-dep)} \nonumber\\
        &+ {\br_i^{(arc-head)}}\tran \bu^{(arc)},
\end{align}
where $\mathrm{MLP}$ is a multi-layer perceptron. A weight
matrix $U^{(arc)}$ determines the strength of a link
from $x_i$ to $x_j$ while $\bu^{(arc)}$ is used in the bias term, which controls the prior headedness of $x_i$.

Thus, $\bs_j^{(arc)} = [s_{1j}^{(arc)}; \cdots;s_{Tj}^{(arc)}]$ is the scores of the potential heads of the $j$-th character, then a softmax function is applied to get the probability distribution.

In the training phase, we minimize the cross-entropy of golden head-dependent pair. In the test phase, we ensure that the resulting parse is a well-formed tree by the heuristics formulated in~\citet{DBLP:conf/iclr/DozatM17}.
%MST algorithm.

\subsubsection{Arc Label Prediction}

After obtaining the best predicted unlabeled tree, we assign the label scores $\bs_{ij}^{(label)}\in \mathbb{R}^K$ for every
arc $x_i \rightarrow x_j$, in which the $k$-th element corresponds to the score of $k$-th label and $K$ is the size of the label set.
In our joint model, the arc label set consists of the standard word-level dependency labels and a special label ``app'' indicating the intra-dependency within a word.

For the arc $x_i \rightarrow x_j$, we obtain $\bs_{ij}^{(label)}$ with
\begin{align}
    \br_i^{(label-head)} &=\mathrm{MLP}^{(label-head)}(\bh_i), \label{eq:label-1}\\
    \br_j^{(label-dep)} &= \mathrm{MLP}^{(label-dep)} (\bh_j), \\
    \br_{ij}^{(label)} &= \br_i^{(label-head)}\oplus \br_j^{(label-dep)},\\
    \bs_{ij}^{(label)} &= \br_i^{(label-head)} \mathcal{U}^{(label)} \br_j^{(label-dep)} \nonumber\\
        &+ W^{(label)}(\br_{ij}^{(label)})  + \bu^{(label)},\label{eq:label-2}
\end{align}
where $\mathcal{U}^{(label)}\in \mathbb{R}^{K\times p\times p}$ is a third-order tensor, $W^{(label)}\in \mathbb{R}^{K\times 2p}$ is a weight matrix, and $\bu^{(label)}\in \mathbb{R}^K$
is a bias vector. The best label of arc $x_i \rightarrow x_j$ is determined according to $\bs_{ij}^{(label)}$.
\begin{align}
y_{ij} = \argmax_{label}\bs_{ij}^{(label)}
\end{align}

In the training phase, we use golden head-dependent relations and cross-entropy to optimize arc label prediction. Characters with continuous ``app'' arcs can be combined into a single word. If a character has no leftward ``app'' arc, it is a single-character word. The arc with label ``app'' is constrained to occur in two adjacent characters and is leftward. When decoding, we first use the proposed model to predict the character-level labeled dependency tree, and then recover the word segmentation and word-level dependency tree based on the predicted character-level arc labels. The characters with continuous ``app'' are regarded as one word. And the predicted head character of the last character is viewed as this word's head. Since the predicted arc points to a character, we regard the word which contains this head character as the head word.

\begin{figure}[t]
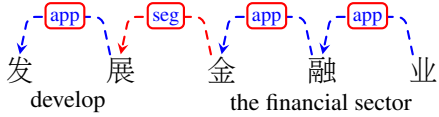

    \centering

\depstyle{app}{edge style = {thick, blue,dashed},
label style = {thick, draw=red, text=blue, fill=white}}
\depstyle{dep}{edge theme = grassy, label style={draw=orange,trapezium}}

\tikzstyle{en}=[font=\footnotesize\selectfont,text=black]
\tikzstyle{mypos}=[font=\footnotesize,text=blue!60!black]
 \begin{dependency}%[theme = simple]
   \begin{deptext}[column sep=2em]
      发 \& 展\& 金\& 融 \& 业  \\
   \end{deptext}

   \node (VV) [en,below of = \wordref{1}{2}, node distance=1em, xshift=-1.8em] {develop};
   \node (VV) [en,below of = \wordref{1}{5}, node distance=1em, xshift=-3.5em] {the financial sector};

   \depedge[app]{2}{1}{app}
   \depedge[app]{4}{3}{app}
   \depedge[app]{5}{4}{app}
   \depedge[app,red]{3}{2}{seg}
\end{dependency}
\caption{Label prediction for word segmentation only. The arc with ``app'' indicates its connected characters belong to a word, and the arc with ``seg'' indicates its connected characters belong to different words.}\label{fig:seg-only}
\end{figure}

\subsection{Models for Word Segmentation Only} \label{Word Segmentation Only}
%duan2018fast.

The proposed model can be also used for Chinese word segmentation task solely.
Without considering the parsing task, we first assign a leftward unlabeled arc by default for every two adjacent characters, and then predict the arc labels which indicate the boundary of segmentation. In the task of word segmentation only, there are two kinds of arc labels: ``seg'' and ``app''. ``seg'' means there is a segmentation between its connected characters, and ``app'' means its connected characters belong to one word. Since the unlabeled arcs are assigned in advance, we just use Eq. \eqref{eq:label-1} $\sim$ \eqref{eq:label-2} to predict the labels: ``seg'' and ``app''. Thus, the word segmentation task is transformed into a binary classification problem.

Fig. \ref{fig:seg-only} gives an illustration of the labeled arcs for the task of word segmentation only.

\section{Experiments} \label{experiments}

\subsection{Datasets}
We use the Penn Chinese Treebank 5.0 (CTB-5)\footnote{\url{https://catalog.ldc.upenn.edu/LDC2005T01}}, 7.0 (CTB-7)\footnote{\url{https://catalog.ldc.upenn.edu/LDC2010T07}} and 9.0 (CTB-9)\footnote{\url{https://catalog.ldc.upenn.edu/LDC2016T13}} datasets to evaluate our models~\cite{DBLP:journals/nle/XueXCP05}. For CTB-5, the training set is from sections 1$\sim$270, 400$\sim$931, and 1001$\sim$1151, the development set is from section 301$\sim$325, and the test set is from section 271$\sim$300, this splitting was also adopted by~\citet{DBLP:conf/emnlp/ZhangC10a,DBLP:conf/acl/ZhangZCL14,DBLP:conf/acl/KuritaKK17}. For CTB-7, we use the same split as~\citet{DBLP:conf/ijcnlp/WangKTCZT11,DBLP:conf/acl/ZhangZCL14,DBLP:conf/acl/KuritaKK17}. For CTB-9, we use the dev and test files proposed by~\citet{DBLP:conf/ijcnlp/ShaoHTN17}, and we regard all left files as the training data.

\subsection{Measures}
 Following~\citet{DBLP:conf/acl/HatoriMMT12,DBLP:conf/acl/ZhangZCL14,DBLP:conf/naacl/ZhangLBD15,DBLP:conf/acl/KuritaKK17}, we use standard measures of word-level F1, precision and recall scores to evaluate word segmentation and dependency parsing (for both unlabeled and labeled scenario) tasks. We detail them in the following.

\begin{itemize*}
  \item $F1_{seg}$: F1 measure of Chinese word segmentation. This is the standard metric used in Chinese word segmentation task~\cite{DBLP:conf/emnlp/QiuZH13,DBLP:conf/ijcai/ChenQH17}.

  \item $F1_{udep}$: F1 measure of unlabeled dependency parsing. Following~\citet{DBLP:conf/acl/HatoriMMT12,DBLP:conf/acl/ZhangZCL14,DBLP:conf/naacl/ZhangLBD15,DBLP:conf/acl/KuritaKK17}, we use standard measures of word-level F1, precision and recall score to evaluate dependency parsing.
      In the scenario of joint word segmentation and dependency parsing, the widely used unlabeled attachment score (UAS) is not enough to measure the performance, since the error arises from two aspects: one is caused by word segmentation and the other is due to the wrong prediction on the head word. A dependent-head pair is correct only when both the dependent and head words are accurately segmented and the dependent word correctly finds its head word.
      The precision of unlabeled dependency parsing (denoted as $P_{udep}$) is calculated by the correct dependent-head pair versus the total number of dependent-head pairs (namely the number of segmented words). The recall of unlabeled dependency parsing (denoted as $R_{udep}$) is computed by the correct dependent-head pair versus the total number of golden dependent-head pairs (namely the number of golden words).
      The calculation of $F1_{udep}$ is like $F1_{seg}$.

  \item $F1_{ldep}$: F1 measure of labeled dependency parsing. The only difference from $F1_{udep}$ is that except the match between the head and dependent words, the pair must have the same label as the golden dependent-head pair.
      The precision and recall are calculated correspondingly. Since the number of golden labeled dependent-head pairs and predicted labeled dependent-head pairs are the same with the counterparts of unlabeled dependency parsing, the value of $F1_{ldep}$ cannot be higher than $F1_{udep}$.

%  \item $UAS$: Unlabeled attachment score. Because in the dependency parsing task, each word has only one head word, the percentage of words that have the correct head can be used as a metric to assess the performance. For our joint dependency parsing, this value equals to the value of the recall of unlabeled dependency parsing ($R_{udep}$).

%  \item $LAS$: Labeled attachment score. This value measures the accuracy of dependency labels, and a dependency label is viewed as correct when its arc and its label are correct. For our unified dependency parsing, this value equals to the value of the recall of labeled dependency parsing ($R_{ldep}$).
\end{itemize*}

The more detailed description of dependency parsing metrics can be found in~\citet{DBLP:series/synthesis/2009Kubler}. The $UAS$, $LAS$ equal to the value of the recall of unlabeled dependency parsing ($R_{udep}$) and the recall of labeled dependency parsing ($R_{ldep}$), respectively. We also report these two values in our experiments.

\subsection{Experimental Settings}
\paragraph{Pre-trained Embedding}
Based on~\citet{DBLP:conf/ijcnlp/ShaoHTN17,DBLP:journals/taslp/ZhangYF18}, n-grams are of great benefit to Chinese word segmentation and POS tagging tasks. Thus we use unigram, bigram and trigram embeddings for all of our character-based models. We first pre-train unigram, bigram and trigram embeddings on Chinese Wikipedia corpus by the method proposed in~\citet{DBLP:conf/naacl/LingDBT15}, which improves standard Word2Vec by incorporating token order information. For a sentence with characters ``abcd...'', the unigram sequence is ``a b c ...''; the bigram sequence is ``ab bc cd ...''; the trigram sequence is ``abc bcd ...''. For our word dependency parser, we use Tencent's pre-trained word embeddings~\cite{DBLP:conf/naacl/SongSLZ18}. Because the Tencent's pre-trained word embedding dimension is 200, we set both pre-trained and random word embedding dimension as 200 for all of our word dependency parsing models. All pre-trained embeddings are fixed during our experiments. In addition to the fixed pre-trained embeddings, we also randomly initialize embeddings, and element-wisely add the pre-trained and random embeddings before other procedures. For model with BERT encoding layer, we use Chinese BERT-base released in~\citet{DBLP:journals/corr/abs-1906-08101}.

\paragraph{Hyper-parameters}
\begin{table}[t]\setlength{\tabcolsep}{8pt}
    \centering\small
    \caption{Hyper-parameter settings.} \label{table:hyper}
    \begin{tabular}{l|c}
      \toprule
      Embedding dimension  & 100 \\
      BiLSTM hidden size & 400  \\
      Gradients clip                      & 5  \\
      Batch size                          & 128 \\
      Embedding dropout                   & 0.33 \\
      LSTM dropout                   & 0.33 \\
      Arc MLP dropout                   & 0.33 \\
      Label MLP dropout                   & 0.33 \\
      LSTM depth                         & 3 \\
      MLP depth                         & 1 \\
      Arc MLP size                   & 500 \\
      Label MLP size                   & 100 \\
      Learning rate                  & 2e-3 \\
      Annealing                     & $.75^{t/5000}$ \\
      $\beta_{1},\beta_{2}$          & 0.9 \\
      Max epochs                  & 100 \\
      \bottomrule
    \end{tabular}

\end{table}

The development set is used for parameter tuning. All random weights are initialized by Xavier normal initializer~\cite{DBLP:journals/jmlr/GlorotB10}.

For BiLSTM based models, we generally follow the hyper-parameters chosen in~\citet{DBLP:conf/iclr/DozatM17}.  The model is trained with the Adam algorithm~\cite{DBLP:journals/corr/KingmaB14} to minimize the sum of the cross-entropy of arc predictions and label predictions. After each training epoch, we test the model on the dev set, and models with the highest $F1_{udep}$ in development set are used to evaluate on the test sets, and the results reported for different datasets in this paper are all on their test set. Detailed hyper-parameters can be found in Table \ref{table:hyper}.

For BERT based models, we use the AdamW optimizer with a  triangle learning rate warmup, the maximum learning rate is $2e-5$ \cite{DBLP:conf/iclr/LoshchilovH19,DBLP:conf/naacl/DevlinCLT19}. It optimizes for 5 epochs, the model has the best development set performance is used to evaluate on the test sets.

\label{compare with previous works}
\begin{table*}[t]\setlength{\tabcolsep}{10pt}
  \centering\small
  \begin{threeparttable}
    \centering
    \begin{tabular}{l*{8}{c}}
      \toprule
      \multirow{2}*{Models} & \multicolumn{2}{c}{CTB-5} & & \multicolumn{2}{c}{CTB-7} & & \multicolumn{2}{c}{CTB-9} \\
      \cmidrule(lr){2-3}\cmidrule(lr){5-6}\cmidrule(lr){8-9}
                                    & $F1_{seg}$ & $F1_{udep}$  &  & $F1_{seg}$ & $F1_{udep}$  &  & $F1_{seg}$ & $F1_{udep}$  \\
      \hline
      \citet{DBLP:conf/acl/HatoriMMT12}  & 97.75 & 81.56 & & 95.42 & 73.58 & & - & -  \\
      \citet{DBLP:conf/acl/ZhangZCL14} STD & 97.67 & 81.63 & & 95.53 & 75.63 & & - & - \\
      \citet{DBLP:conf/acl/ZhangZCL14} EAG & 97.76 & 81.70 & & 95.39 & 75.56 & & - & - \\
      \citet{DBLP:conf/naacl/ZhangLBD15}    & 98.04 & 82.01 & & -     & -     & & - & - \\
      \citet{DBLP:conf/acl/KuritaKK17}       & 98.37 & 81.42 & & 95.86 & 74.04 & & - & - \\
      \midrule
      Joint-Binary          & 98.45 & 87.24 & & 96.57 & 81.34 & & 97.10 & 81.67  \\
      %\midrule
      Joint-Multi           & \textbf{98.48} & \textbf{87.86} & & \textbf{96.64} & \textbf{81.80} & & \textbf{97.20} & \textbf{82.15}  \\
      %\midrule
      Joint-Multi-BERT            & 98.46 & \textbf{89.59} & & \textbf{97.06} & \textbf{85.06} & & \textbf{97.63} & \textbf{85.66} \\
      \bottomrule
    \end{tabular}
      \begin{tablenotes}
\item [1] STD and EAG in
  ~\citet{DBLP:conf/acl/ZhangZCL14} denote the arc-standard and the arc-eager models.
\item [2] $F1_{seg}$ and $F1_{udep}$ are the F1 score for Chinese word segmentation and unlabeled dependency parsing, respectively.
    %More details on our models and metrics can be found in Section \ref{experiments} .
\end{tablenotes}
    \caption{Main results in the test set of different datasets. Our Joint-Multi model achieves superior performance than previous joint models. The Joint-Multi-BERT further enhances the performance of dependency parsing significantly.} \label{tb:exp1}
    \end{threeparttable}
\end{table*}

\subsection{Proposed Models}\label{sec:exp-prop-model}
In this part, we introduce the settings for our proposed joint models. Based on the way the model uses dependency parsing labels and encoding layers, we divide our models into four kinds. We enumerate them as follows.
\begin{itemize}
  \item \textbf{Joint-SegOnly model}:
    The proposed model can be also used for word segmentation task only. In this scenario, the dependency arcs are just allowed to appear in two adjacent characters and $label\in\{app,seg\}$. This model is described in Section \ref{Word Segmentation Only}.
  \item \textbf{Joint-Binary model}:
    This scenario means $label\in\{app, dep\}$. In this situation, the label information of all the dependency arcs is ignored. Each word-level dependency arc is labeled as $dep$, the intra-word dependency is regarded as $app$. Characters with continuous $app$ label will be joined together and viewed as one word. The $dep$ label indicates this character is the end of a word.
  \item \textbf{Joint-Multi model}:
    This scenario means $label\in\{app,dep_1,\cdots,dep_K\}$, where $K$ is the number of types of dependency arcs. The intra-word dependency is viewed as $app$. The other labels are the same as the original arc labels. But instead of representing the relationship between two words, the labeled arc represents the relationship between the last character of the dependent word and the last character of the head word.
  \item \textbf{Joint-Multi-BERT model}:
    For this kind of models, the encoding layer is BERT. And it uses the same target scenario as the Joint-Multi model.
\end{itemize}

\subsection{Comparison with the Previous Joint Models}

In this part, we mainly focus on the performance comparison between our proposed models and the previous joint models. Since the previous models just deal with the unlabeled dependency parsing, we just report the $F1_{seg}$ and $F1_{udep}$ here.

As presented in Table \ref{tb:exp1}, our model (\textbf{Joint-Binary}) outpaces previous methods with a large margin in both Chinese word segmentation and dependency parsing, even without the local parsing features which were extensively used in previous transition-based joint work~\cite{DBLP:conf/acl/HatoriMMT12,DBLP:conf/acl/ZhangZCL14,DBLP:conf/naacl/ZhangLBD15,DBLP:conf/acl/KuritaKK17}.
Another difference between our joint models and previous works is the combination of POS tags, the previous models all used the POS task as one componential task. Despite the lack of POS tag information, our models still achieve much better results. However, according to~\citet{DBLP:conf/iclr/DozatM17}, POS tags are beneficial to dependency parsers, therefore one promising direction of our joint model might be incorporating POS tasks into this joint model.

Other than the performance distinction between previous work, our joint model with or without dependency labels also differ from each other. It is clearly shown in Table \ref{tb:exp1}, our joint model with labeled dependency parsing (\textbf{Joint-Multi}) outperforms its counterpart (\textbf{Joint-Binary}) in both Chinese word segmentation and dependency parsing. With respect to the enhancement of dependency parsing caused by the arc labels, we believe it can be credited to two aspects. The first one is the more accurate Chinese word segmentation. The second one is that label information between two characters will give extra supervision for the search of head characters.
The reason why labeled dependency parsing is conducive to the Chinese word segmentation will be also analyzed in Section \ref{Chinese Word Segmentation}.

Owing to the joint decoding of Chinese word segmentation and dependency parsing, we can utilize the character-level pre-trained language model BERT. The last row of Table \ref{tb:exp1} displays the $F1_{udep}$ can be substantially increased when BERT is used, even when the performance of Chinese word segmentation does not improve too much. We presume this indicates BERT can better extract the contextualized information to help the dependency parsing.

\subsection{Chinese Word Segmentation} \label{Chinese Word Segmentation}

\begin{table*}[t]\setlength{\tabcolsep}{2pt}
  \centering\small
  \begin{threeparttable}
    \begin{tabular}{lc*{12}{c}}
      \hline
      \multirow{2}*{Models} & \multirow{2}*{Tag Set} & \multicolumn{3}{c}{CTB-5} & & \multicolumn{3}{c}{CTB-7} & & \multicolumn{3}{c}{CTB-9} \\
                          \cmidrule(lr){3-5}\cmidrule(lr){7-9} \cmidrule(lr){11-13}
                 & & $F1_{seg}$ & $P_{seg}$ & $R_{seg}$ & & $F1_{seg}$ & $P_{seg}$ & $R_{seg}$ & & $F1_{seg}$ & $P_{seg}$ & $R_{seg}$  \\
      \midrule
      LSTM+MLP & $\{B,M,E,S\}$   & 98.47 & 98.26 & 98.69 &  & 95.45 & 96.44 & 96.45  &  & 97.11 & 97.19 & 97.04  \\
      LSTM+CRF & $\{B,M,E,S\}$   & 98.48 & 98.33 & 98.63 &  & 96.46 & 96.45 & 96.47  &  & 97.15 & 97.18 & 97.12   \\
      LSTM+MLP & $\{app,seg\}$ & 98.40 & 98.14 & 98.66 & & 96.41 & 96.53 & 96.29 &  & 97.09 & 97.16 & 97.02 \\
      \midrule
      Joint-SegOnly & $\{app,seg\}$ & \textbf{98.50} & \textbf{98.30} & 98.71 &  & 96.50 & 96.67 & 96.34  &  & 97.09 & 97.15 & 97.04   \\
      Joint-Binary & $\{app, dep\}$ & 98.45 & 98.16 & 98.74 & & 96.57 & 96.66 & 96.49  &  & 97.10 & 97.16 & 97.04  \\
      Joint-Multi & $\{app,dep_1,\cdots,dep_K\}$ & 98.48 & 98.17 & \textbf{98.80} & & \textbf{96.64} & \textbf{96.68} & \textbf{96.60}  &  & \textbf{97.20} & \textbf{97.31} & \textbf{97.19}  \\
      Joint-Multi-BERT & $\{app,dep_1,\cdots,dep_K\}$ & 98.46 & 98.12 & \textbf{98.81} & & \textbf{97.06} & \textbf{97.05} & \textbf{97.08} & & \textbf{97.63} & \textbf{97.68} & \textbf{97.58} \\
      \bottomrule
    \end{tabular}
          \begin{tablenotes}
\item [1] The upper part refers the models based on sequence labeling.
\item [2] The lower part refers our proposed joint models which are detailed in section \ref{sec:exp-prop-model}. The proposed joint models achieve near or better $F1_{seg}$ than models trained only on Chinese word segmentation.
\item [3] $F1_{seg}$, $P_{seg}$ and $R_{seg}$ are the F1, precision and recall of Chinese word segmentation, respectively.
\end{tablenotes}
  \caption{Results of Chinese word segmentation.}\label{tb:exp2}

  \end{threeparttable}
\end{table*}
\begin{table*}[t]\setlength{\tabcolsep}{1pt}
  \centering\footnotesize
  \begin{threeparttable}
   \begin{tabular}{l*{15}{c}}
      \toprule
      \multirow{2}*{Models}
      & \multicolumn{5}{c}{CTB-5} & \multicolumn{5}{c}{CTB-7} & \multicolumn{5}{c}{CTB-9}\\
      \cmidrule(lr){2-6}\cmidrule(lr){7-11}\cmidrule(lr){12-16}
                & $F1_{seg}$ & $F1_{udep}$ & $UAS$ & $F1_{ldep}$ & $LAS$ &  $F1_{seg}$ & $F1_{udep}$ & $UAS$ & $F1_{ldep}$ & $LAS$ & $F1_{seg}$ & $F1_{udep}$ & $UAS$ & $F1_{ldep}$ & $LAS$\\
      \hline
      Biaffine\tnote{$\dagger$}  & -   & - & 88.81 & - & 85.63 & -   & - & 86.06 & - & 81.33 & -   & - & 86.21 & - & 81.57 \\
      \midrule
      Pipeline\tnote{$\mathsection$}  & \textbf{98.50}  & 86.50 & 86.71 & 83.46 & 83.67 &  96.50 & 80.62 & 80.49 & 76.58 & 76.46 & 97.09  & 81.54 & 81.61 & 77.34 & 77.40 \\
      Joint-Multi     & 98.48 & \textbf{87.86} & \textbf{88.08} & \textbf{85.08} & \textbf{85.23} &  \textbf{96.64} & \textbf{81.80} & \textbf{81.80} & \textbf{77.84} & \textbf{77.83} & \textbf{97.20} & \textbf{82.15} & \textbf{82.23} & \textbf{78.08} & \textbf{78.14}\\
      Joint-Multi-BERT   &  98.46  & \textbf{89.59} & \textbf{89.97} & \textbf{85.94} & \textbf{86.3} & \textbf{97.06} & \textbf{85.06}  & \textbf{85.12} & \textbf{80.71} & \textbf{80.76} & \textbf{97.63} & \textbf{85.66} & \textbf{85.74} & \textbf{81.71} & \textbf{81.77} \\
      \bottomrule
    % \end{tabular}
    \end{tabular}
     \begin{tablenotes}
\item[$\dagger$] The results are evaluated by a word-level biaffine parser on the gold-segmented sentences.
\item[$\mathsection$] The pipeline model first uses the Joint-SegOnly model to segment the sentence, then uses the word-level biaffine parser to get the parsing result.
\end{tablenotes}
\caption{Comparison with the pipeline model. Our Joint-Multi models outperform the pipeline models in a large margin. When BERT is used, the dependency parsing performance was significantly improved, although the Chinese word segmentation does not meliorate a lot.} \label{tb:exp3}

\end{threeparttable}

\end{table*}

In this part, we focus on the performance of our model for the Chinese word segmentation task only.

Since most of state-of-the-art Chinese word segmentation methods are based on sequence labeling, in which every sentence is transformed into a sequence of $\{B,M,E,S\}$ tags. $B$ represents the begin of a word, $M$ represents the middle of a word, $E$ represents the end of a word, $S$ represents the word has only one character.
We compare our model with these state-of-the-art methods.

\begin{itemize*}
  \item LSTM+MLP with $\{B,M,E,S\}$ tags.
    Following~\citet{DBLP:conf/emnlp/MaGW18}, we tried to do Chinese word segmentation without conditional random field (CRF). After BiLSTM, the hidden states of each character further forwards into a multi layer perceptron (MLP), so that every character can output a probability distribution over the label set. Viterbi algorithm is utilized to find the global maximum label sequence when testing.
  \item LSTM+CRF with $\{B,M,E,S\}$ tags.
    The only difference between this scenario and the previous one is whether using conditional random field (CRF) after the MLP~\cite{DBLP:conf/icml/LaffertyMP01,DBLP:conf/ijcai/ChenQH17}.
  \item LSTM+MLP with $\{app,seg\}$ tags.
    The segmentation of a Chinese sentence can be represented by a sequence of $\{app,seg\}$, where $app$ represents the next character and this character belongs to the same word, and $seg$ represents this character is the last character of a word. Therefore, Chinese word segmentation can be viewed as a binary classification problem. Except for the tag set, this model's architecture is similar to the LSTM+MLP scenario.
\end{itemize*}

All the above models use the multi-layer BiLSTM as the encoder, they differ from each other in their way of decoding and the tag set. The number of BiLSTM layers is 3 and the hidden size is 200.

The performance of all models are listed in Table \ref{tb:exp2}. The first two rows present the difference between whether utilizing CRF on the top of MLP. CRF's performance is slightly better than its counterpart. The first row and the third row display the comparison between different tag scenario, the $\{B,M,E,S\}$ tag set is slightly better than the $\{app,seg\}$ tag set.

Different to the competitor sequence labeling model (LSTM+MLP with $\{app,seg\}$ tag set), our joint-SegOnly model uses the biaffine to model the interaction between the two adjacent characters near the boundary and achieves slightly better or similar performances on all datasets. The empirical results in three datasets suggest that modeling the interaction between two consecutive characters are helpful to Chinese word segmentation. If two characters are of high probability to be in a certain dependency parsing relationship, there will be a greater chance that one of the characters is the head character.

The lower part of Table \ref{tb:exp2} shows the segmentation evaluation of the proposed joint models. Jointly training Chinese word segmentation and dependency parsing achieves comparable or slightly better Chinese word segmentation than training Chinese word segmentation alone. Although head prediction is not directly related to Chinese word segmentation, the head character can only be the end of a word, therefore combination between Chinese word segmentation and character dependency parsing actually introduces more supervision for the former task. On CTB-5, the Joint-binary and Joint-multi models are slightly worse than the joint-SegOnly model. The reason may be that the CTB-5 dataset is relatively small and the complicated models suffer from the overfitting. From the last row of Table \ref{tb:exp2}, BERT can further enhance the model's performance on Chinese word segmentation.

\begin{table*}[t]\setlength{\tabcolsep}{1pt}
  \centering\footnotesize
  \begin{threeparttable}
    \begin{tabular}{l*{16}{c}}
      \toprule
       \multirow{2}*{Models}& \multicolumn{5}{c}{CTB-5} &  \multicolumn{5}{c}{CTB-7}& \multicolumn{5}{c}{CTB-9}
      \\
      \cmidrule(lr){2-6}\cmidrule(lr){7-11}\cmidrule(lr){12-16}
      & $F1_{seg}$ & $F1_{udep}$ & $UAS$ & $F1_{ldep}$ & $LAS$  & $F1_{seg}$ & $F1_{udep}$ & $UAS$ & $F1_{ldep}$ & $LAS$ &   $F1_{seg}$ & $F1_{udep}$ & $UAS$ & $F1_{ldep}$ & $LAS$ \\

      \midrule
      Joint-Multi  & 98.48 & 87.86 & 88.08 & 85.08 & 85.23 &  96.64 & 81.80 & 81.80 & 77.84 & 77.83 & 97.20 & 82.15 & 82.23 & 78.08 & 78.14 \\
      \quad -pre-trained & 97.72 & 82.56 & 82.70 & 79.8 & 70.93 &  95.52 & 76.35 & 76.22 & 72.16 & 72.04  & 96.56 & 78.93 & 78.93 & 74.35 & 74.37\\
      \quad -n-gram & 97.72 & 83.44 & 83.60 & 80.24 & 80.41 &  95.21 & 77.37 & 77.11 & 72.94 & 72.69 & 95.85 & 78.55 & 78.41 & 73.94 & 73.81\\
      \bottomrule

    \end{tabular}
    \begin{tablenotes}
    \item [1] The `-pre-trained' means the model is trained without the pre-trained embeddings.
    \item [2] The `-n-gram' means the model is trained by removing the bigram and trigram embeddings, only randomly initialized and pre-trained character embeddings are used.
\end{tablenotes}
    \caption{Ablation experiments for Joint-Multi models. } \label{tb:ablation}

    \end{threeparttable}
\end{table*}

\begin{table*}[ht]\setlength{\tabcolsep}{1pt}
  \centering\footnotesize
  \begin{threeparttable}
    \centering
    \begin{tabular}{l*{11}{c}}
      \toprule
      \multirow{2}*{Models} & \multicolumn{3}{c}{CTB-5} & & \multicolumn{3}{c}{CTB-7}  & & \multicolumn{3}{c}{CTB-9}\\
          \cmidrule(lr){2-4}\cmidrule(lr){6-8}\cmidrule(lr){10-12}
            & $P_{udep}$ & Seg-wrong & Head-wrong & & $P_{udep}$ & Seg-wrong & Head-wrong & & $P_{udep}$ & Seg-wrong & Head-wrong \\
      \hline
       Pipeline   & 86.28\%  & 3.49\% & 10.23\% & & 80.75\% & 7.10\% & 12.15\% & & 81.48\% & 6.76\% & 11.76\%  \\
       Joint-Multi & 87.65\% & 3.43\% & 8.92\% &  & 81.81\% & 6.80\% & 11.39\% & & 82.08\% & 6.55\% & 11.37\% \\
       Joint-Multi-BERT & 89.22\% & 3.2\% & 7.58\% & & 85.01\% & 6.10\% & 8.89\% & & 85.59\% & 5.74\% & 8.89\% \\
      \bottomrule
    \end{tabular}
        \begin{tablenotes}
\item [1] The value of $P_{udep}$ is the percentage that the predicted arc is correct. `Seg-wrong' means that either head or dependent (or both) is wrongly segmented. `Head-wrong' means that the word is correctly segmented but the predicted head word is wrong.
\end{tablenotes}
    \caption{Error analysis of unlabeled dependency parsing in the test set of different datasets.}\label{tb:exp4}

    \end{threeparttable}
\end{table*}

\begin{figure*}[t]
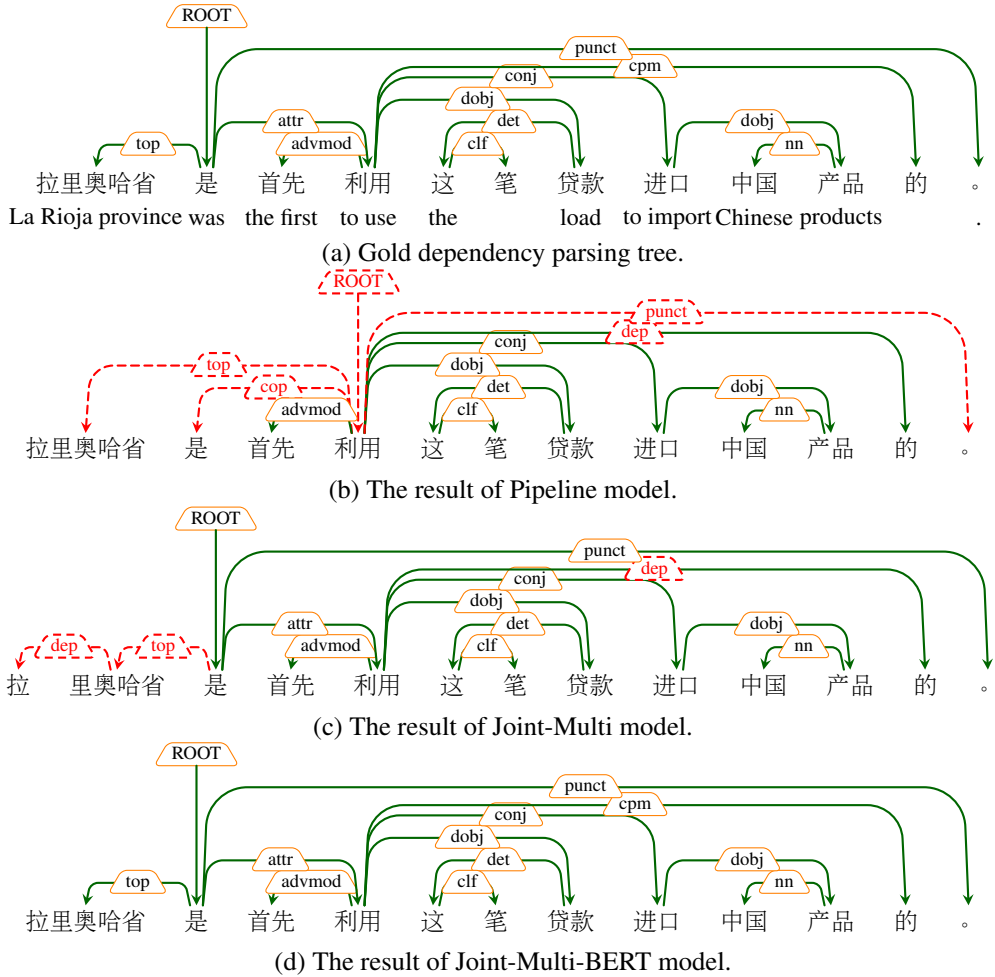

    \centering\small
\depstyle{dep}{edge theme = grassy, label style={draw=orange,trapezium}}
\depstyle{dep_label_wrong}{edge theme = grassy,
            label style = {thick, draw=red, densely dashed, text=red, trapezium}}
\depstyle{dep_wrong}{edge style={thick,red,densely dashed}, label style = {thick, text=red, draw=red, densely dashed, trapezium}}

\tikzstyle{en}=[text height=0,anchor=north,font=\footnotesize\selectfont,text=black,inner sep=0]
\tikzstyle{mypos}=[font=\footnotesize,text=blue!60!black]

\begin{dependency}[scale=0.6]
  \begin{deptext}[column sep=1em,en]
    拉里奥哈省 \& 是 \& 首先 \& 利用 \& 这 \& 笔 \& 贷款 \& 进口 \& 中国 \& 产品 \& 的 \& 。  \\
  \end{deptext}

  \node (shanghai) [en,below = of \wordref{1}{1}, node distance=0.2em, yshift=2em] {La Rioja province};
  \node (plan) [en,below  = of \wordref{1}{2}, node distance=1em, yshift=2em] {was};
  \node (VV) [en,below = of \wordref{1}{3}, node distance=1em, yshift=2em] {the first};
  \node (VV) [en,below = of \wordref{1}{4}, node distance=1em, yshift=2em] {to use};
  \node (VV) [en,below = of \wordref{1}{5}, node distance=1em, yshift=2em] {the};
  \node (VV) [en,below = of \wordref{1}{7}, node distance=1em, yshift=2em] {load};
  \node (VV) [en,below = of \wordref{1}{8}, node distance=1em, yshift=2em] {to import};
  \node (VV) [en,below = of \wordref{1}{9}, node distance=1em, yshift=2em] {Chinese};
  \node (VV) [en,below = of \wordref{1}{10}, node distance=1em, yshift=2em] {products};
  \node (VV) [en,below = of \wordref{1}{12}, node distance=1em, yshift=2em] {.};

  \deproot[dep,edge unit distance=3.5ex]{2}{ROOT}
  \depedge[dep]{2}{1}{top}
  \depedge[dep]{4}{3}{advmod}
  \depedge[dep]{2}{4}{attr}
  \depedge[dep]{7}{5}{det}
  \depedge[dep]{5}{6}{clf}
  \depedge[dep]{4}{7}{dobj}
  \depedge[dep]{4}{8}{conj}
  \depedge[dep]{10}{9}{nn}
  \depedge[dep]{8}{10}{dobj}
  \depedge[dep,edge unit distance=2ex]{4}{11}{cpm}
  \depedge[dep,edge unit distance=1.65ex]{2}{12}{punct}
\end{dependency}

(a) Gold dependency parsing tree.

\begin{dependency}[scale=0.6]
  \begin{deptext}[column sep=1em,en]
    拉里奥哈省 \& 是 \& 首先 \& 利用 \& 这 \& 笔 \& 贷款 \& 进口 \& 中国 \& 产品 \& 的 \& 。  \\
  \end{deptext}
  \deproot[dep_wrong,edge unit distance=3.5ex]{4}{ROOT}
  \depedge[dep_wrong]{4}{1}{top}
  \depedge[dep_wrong]{4}{2}{cop}
  \depedge[dep]{4}{3}{advmod}
  \depedge[dep]{7}{5}{det}
  \depedge[dep]{5}{6}{clf}
  \depedge[dep]{4}{7}{dobj}
  \depedge[dep]{4}{8}{conj}
  \depedge[dep]{10}{9}{nn}
  \depedge[dep]{8}{10}{dobj}
  \depedge[dep_label_wrong,edge unit distance=2ex]{4}{11}{dep}
  \depedge[dep_wrong,edge unit distance=2.1ex]{4}{12}{punct}
\end{dependency}

(b) The result of Pipeline model.

\begin{dependency}[scale=0.6]
  \begin{deptext}[column sep=1em,en]
    拉 \& 里奥哈省 \& 是 \& 首先 \& 利用 \& 这 \& 笔 \& 贷款 \& 进口 \& 中国 \& 产品 \& 的 \& 。  \\
  \end{deptext}
  \deproot[dep,edge unit distance=3.5ex]{3}{ROOT}
  \depedge[dep_wrong]{2}{1}{dep}
  \depedge[dep_wrong]{3}{2}{top}
  \depedge[dep]{5}{4}{advmod}
  \depedge[dep]{3}{5}{attr}
  \depedge[dep]{8}{6}{det}
  \depedge[dep]{6}{7}{clf}
  \depedge[dep]{5}{8}{dobj}
  \depedge[dep]{5}{9}{conj}
  \depedge[dep]{11}{10}{nn}
  \depedge[dep]{9}{11}{dobj}
  \depedge[dep_label_wrong,edge unit distance=2ex]{5}{12}{dep}
  \depedge[dep,edge unit distance=1.65ex]{3}{13}{punct}
\end{dependency}

(c) The result of Joint-Multi model.

\begin{dependency}[scale=0.6]
  \begin{deptext}[column sep=1em,en]
    拉里奥哈省 \& 是 \& 首先 \& 利用 \& 这 \& 笔 \& 贷款 \& 进口 \& 中国 \& 产品 \& 的 \& 。  \\
  \end{deptext}
  \deproot[dep,edge unit distance=3.5ex]{2}{ROOT}
  \depedge[dep]{2}{1}{top}
  \depedge[dep]{4}{3}{advmod}
  \depedge[dep]{2}{4}{attr}
  \depedge[dep]{7}{5}{det}
  \depedge[dep]{5}{6}{clf}
  \depedge[dep]{4}{7}{dobj}
  \depedge[dep]{4}{8}{conj}
  \depedge[dep]{10}{9}{nn}
  \depedge[dep]{8}{10}{dobj}
  \depedge[dep,edge unit distance=2ex]{4}{11}{cpm}
  \depedge[dep,edge unit distance=1.65ex]{2}{12}{punct}
\end{dependency}

(d) The result of Joint-Multi-BERT model.

\caption{Parsing results of different models. The red dashed box means the dependency label is wrong. The red dashed edge means this dependency arc does not exist. Although the Pipeline model has the right word segmentation, ``拉里奥哈省'' is an OOV. Therefore, it fails to find the right dependency relation and adversely affects predictions afterward. The Joint-Multi model can still have a satisfying outcome even with wrong segmentation, which depicts that the Joint-Multi model is resistant to wrong word segmentations. The Joint-Multi-BERT correctly finds the word segmentation and dependency parsing.}\label{fig:error-parsing} \label{fig:examples}
\end{figure*}

Another noticeable phenomenon from the lower part of Table \ref{tb:exp2} is that the labeled dependency parsing brings benefit to Chinese word segmentation. We assume this is because the extra supervision from dependency parsing labels is informative for word segmentation.

% HERE
\subsection{Comparison with the Pipeline Model}

In this part, we compare our joint model with the pipeline model. The pipeline model first uses our best Joint-SegOnly model to get segmentation results, then apply the word-based biaffine parser to parse the segmented sentence.  The word-level biaffine parser is the same with~\citet{DBLP:conf/iclr/DozatM17} but without POS tags. Just like the joint parsing metric, for a dependent-head word pair, only when both head and dependent words are correct, this pair can be viewed as a right one.

From Table \ref{tb:exp3}, it obviously shows that in CTB-5, CTB-7 and CTB-9, the Joint-Multi model consistently outperforms the pipeline model in $F1_{udep}$, $UAS$, $F1_{ldep}$ and $LAS$.
Although the $F1_{seg}$ difference between the Joint-Multi model and the pipeline model is only -0.03, +0.15, +0.11 in CTB-5, CTB-7 and CTB-9, respectively, the $F1_{udep}$ of the Joint-Multi is higher than the pipeline model by +1.36, +1.18 and +0.61 respectively, we believe this indicates the better resistance to error propagation of the Joint-Multi model.

Besides, when BERT is used, $F1_{udep}$, $UAS$, $F1_{ldep}$, and $LAS$ are substantially improved, which represents that our joint model can take advantage of the power of BERT. In CTB-5, the joint model even achieves better $UAS$ than the gold-segmented word-based model. And for the $LAS$, Joint-Multi-BERT models also achieve better results in CTB-5 and CTB-9. We presume the reason that performance of  Chinese word segmentation does not improve as much as dependency parsing is the falsely segmented words in Joint-Multi-BERT are mainly segmenting a long word into several short words or recognizing several short words as one long word.

\subsection{Ablation Study}

As our model uses various n-gram pre-trained embeddings, we explore the influence of these pre-trained embeddings. The second row in Table \ref{tb:ablation} shows the results of the Joint-Multi model without pre-trained embeddings, it is clear that pre-trained embeddings are important for both the word segmentation and dependency parsing.

We also tried to remove the bigram and trigram. Results are illustrated in the third row of Table \ref{tb:ablation}. Compared with the Joint-Multi model, without bigram and trigram, it performs worse in all metrics. However, the comparison between the second row and the third row shows divergence in Chinese word segmentation and dependency parsing for dataset CTB-5 and CTB-7. For Chinese word segmentation, the model without pre-trained embeddings gets superior performance than without bigram and trigram feature. While for all dependency parsing related metrics, the model with pre-trained character embedding gets better performance. We assume the n-gram features are important to Chinese word segmentation. But for the dependency parsing task, the relation between two characters are more beneficial, when pre-trained embeddings are combined, the model can exploit the relationship encoded in the pre-trained embeddings. Additionally, for CTB-5 and CTB-7, even though the third row has inferior $F1_{seg}$ (in average 0.16\% lower than the second row), it still achieves much better $F1_{udep}$ (in average 0.95\% higher than the second row). We believe this is a proof that joint Chinese word segmentation and dependency parsing is resistant to error propagation. The higher segmentation and dependency parsing performance for the model without pre-trained embedding in CTB-9 might owing to its large training set, which can achieve better results even from randomly initialized embeddings.

\subsection{Error Analysis}

Apart from performing the standard evaluation, we investigate where the dependency parsing head prediction error comes from. The errors can be divided into two kinds, one is either the head or dependent (or both) wrongly segmented, the other is the wrong choice on the head word. The ratio of these two mistakes is presented in Table \ref{tb:exp4}. For the Joint-Multi model, more mistakes caused by segmentation in CTB-7 is coherent with our observation that CTB-7 bears lower Chinese word segmentation performance. Based on our error analysis, the wrong prediction of head word accounts for most of the errors, therefore further joint models address head prediction error problem might get more gain on performance.

Additionally, although from Table \ref{tb:exp3} the distinction of $F1_{seg}$ between the Joint-Multi model and the Pipeline model is around +0.1\% in average, the difference between the Head-wrong is more than around +0.82\% in average. We think this is caused by the Pipeline model is more sensitive to word segmentation errors and suffers more from the OOV problem, as depicted in Fig. \ref{fig:examples}. From the last row of Table \ref{tb:exp3}, Joint-Multi-BERT achieves excellent performance on dependency parsing because it can significantly reduce errors caused by predicting the wrong head.

\section{Conclusion and Future Work}
In this paper, we propose a graph-based model for joint Chinese word segmentation and dependency parsing. Different from the previous joint models, our proposed model is a graph-based model and more concise, which results in fewer efforts of feature engineering. Although no explicit handcrafted parsing features are applied, our joint model outperforms the previous feature-riched joint models in a large margin. The empirical results in CTB-5, CTB-7 and CTB-9 show that the dependency parsing task is also beneficial to Chinese word segmentation. Besides, labeled dependency parsing not only is good for Chinese word segmentation, but also avails the dependency parsing head prediction.

Apart from the good performance, the comparison between our joint model and the pipeline model shows great potentialities for character-based Chinese dependency parsing. And owing to the joint decoding between Chinese word segmentation and dependency parsing, our model can use the pre-trained character-level language model (such as BERT) to enhance the performance further. After the incorporation of BERT, the performance of our joint model increases substantially, making the character-based dependency parsing even performs near the gold-segmented word-based dependency parsing. Our proposed method not merely outpaces the pipeline model, but also avoids the preparation for pre-trained word embeddings which depends on a good Chinese word segmentation model.

In order to fully explore the possibility of graph-based Chinese dependency parsing, future work should be done to incorporate the POS tagging into this framework. Additionally, as illustrated in~\citet{DBLP:conf/acl/ZhangZCL14}, more reasonable intra-word dependent structure might further boost the performance of all tasks.

\section*{Acknowledgements}
We would like to thank the action editor and the anonymous reviewers for their insightful comments. We also thank the developers of fastNLP\footnote{\url{https://github.com/fastnlp/fastNLP}}, Yunfan Shao and Yining Zheng, to develop this handy natural language processing package. This work was supported by the National Key Research and Development Program of China (No. 2018YFC0831103), National Natural Science Foundation of China (No. 61672162), Shanghai Municipal Science and Technology Major Project (No. 2018SHZDZX01) and ZJLab.
\end{CJK*}

% \bibliography{naaclhlt2019,nlp,joint}
\bibliography{joint}
\bibliographystyle{acl_natbib}

\end{document}